\documentclass{myifcolog}

\usepackage[utf8]{inputenc} 
\usepackage[T1]{fontenc}    
\usepackage{times}
\usepackage{url}            
\usepackage{booktabs}       
\usepackage{amsfonts}       
\usepackage{nicefrac}       
\usepackage{microtype}      
\usepackage{amssymb}
\usepackage{ragged2e}
\usepackage[toc,page]{appendix}
\usepackage{tikz}
\usetikzlibrary{arrows,positioning,shapes.misc,shapes.geometric}
\usepackage{xcolor}
\usepackage{xspace}
\usepackage[]{todonotes}
\usepackage{bbm}
\usepackage{color,soul}
\usepackage{wrapfig}
\usepackage{subfig}
\usepackage{authblk}

\jvolume{6}
\jnumber{4}
\jyear{2019}

\newcommand{\pphi}{\hat{p}(\phi|\bo, \btheta)}
\newcommand{\pnpsi}{\hat{p}(\neg \psi|\bo, \btheta)}
\newcommand{\ppsi}{\hat{p}(\psi|\bo, \btheta)}
\newcommand{\pnphi}{\hat{p}(\neg \phi|\bo, \btheta)}

\newcommand{\pred}[1]{\textsf{#1}}
\newcommand{\fp}[1]{f_{\pred{#1}}^{\btheta}}

\newcommand{\predP}{\textsf{P}}
\newcommand{\bx}{\mathbf{x}}
\newcommand{\btheta}{\boldsymbol{\theta}}

\newcommand{\bo}{\mathbf{o}}
\newcommand{\bw}{\mathbf{w}}
\newcommand{\bA}{\mathbf{A}}

\newcommand{\dmp}{d_{\varphi}^{\mathrm{MP}}}
\newcommand{\dmt}{d_{\varphi}^{\mathrm{MT}}}

\DeclareMathOperator*{\argmax}{arg\,max}
\DeclareMathOperator*{\argmin}{arg\,min}

\newcommand{\prl}{\hat{p}}

\newcommand{\corpus}{\mathcal{K}}
\newcommand{\loss}{\mathcal{L}}
\newcommand{\crmp}{\mathrm{cr}^{\mathrm{MP}}}

\newcommand{\cump}{\mathrm{cu}^{\mathrm{MP}}}

\title{Semi-Supervised Learning using Differentiable Reasoning}
\titlerunning{Differentiable Reasoning}
\titlethanks{We thank the anonymous reviewers of NeSy for their valueable feedback. Additionally we thank Luciano Serafini, Jasper Driessens, Fije Overeem and also the other attendees of NeSy for the great discussions.}

\date{}

\addauthor[e.van.krieken@vu.nl]
   {Emile van Krieken}
   {VU University Amsterdam}

\addauthor[erman.acar@vu.nl]
   {Erman Acar}
   {VU University Amsterdam}

\addauthor[Frank.van.Harmelen@vu.nl]
   {Frank van Harmelen}
   {VU University Amsterdam}
   
\authorrunning{van Krieken, Acar and van Harmelen}

\begin{document}
\theoremstyle{definition}
\newtheorem{exmp}{Example}
\maketitle

\begin{abstract}
We introduce Differentiable Reasoning (DR), a novel semi-supervised learning technique which uses relational background knowledge to benefit from unlabeled data. We apply it to the Semantic Image Interpretation (SII) task and show that background knowledge provides significant improvement. We find that there is a strong but interesting imbalance between the contributions of updates from Modus Ponens (MP) and its logical equivalent Modus Tollens (MT) to the learning process, suggesting that our approach is very sensitive to a phenomenon called the Raven Paradox \cite{hempel1945studies}. We propose a solution to overcome this situation.  
\end{abstract}

\section{Introduction}
Semi-supervised learning is a common class of methods for machine learning tasks where we consider not just labeled data, but also make use of unlabeled data \cite{chapelle2009semi}. This can be very beneficial for training in tasks where labeled data is much harder to acquire than unlabeled data. 

One such task is \textit{Semantic Image Interpretation (SII)} in which the goal is to generate a semantic description of the objects on an image \cite{donadello2017logic}. This description is represented as a labeled directed graph, which is known as a \textit{scene graph} \cite{johnson2015image}. 
 An example of a labeled dataset for this problem is VisualGenome \cite{krishna2017visual} which contains 108,077 images to train 156,722 different unary and binary predicates. The binary relations in particular make this dataset very sparse, as there are many different pairs of objects that could be related. However, a far larger, though unfortunately unlabeled, dataset like ImageNet \cite{ILSVRC15} contains over 14 million different pictures. Because it is so much larger, it will have many examples of interactions that are not present in VisualGenome. We show that it is possible to improve the performance of a simple classifier on the SII task significantly by adding the satisfaction of a first-order logic (FOL) knowledge base to the supervised loss function. The computation of this satisfaction uses an unlabeled dataset as its domain.

For this purpose, we introduce a statistical relational learning framework called \textit{Differentiable Reasoning} (DR) in Section \ref{diff_reasoning}, as our primary contribution. DR uses simple logical formulas to deduce new training examples in an unlabeled dataset. This is done by adding a differentiable loss term that evaluates the truth value of the formulas. 

In the experimental analysis, we find that the gradient updates using the Modus Ponens (MP) and Modus Tollens (MT) rules are disproportionate. That is, MT often strongly dominates MP in the learning process. Such behavior suggests that our approach is highly sensitive to the Raven Paradox \cite{hempel1945studies}. It refers to the phenomenon that the observations obtained from ``All ravens are black'' are dominated by its logically equivalent ``All non-black things are non-ravens''. Indeed, this is closely related to the \emph{material implication} which caused a lot of discussion throughout the history of logic and philosophy \cite{sep-conditionals}. Our second main contribution relies on its investigation in Section \ref{sec:raven}, and our proposal to cope with it. Finally, we show results on a simple dataset in Section \ref{sec:experiments} and analyze the behavior of the Raven Paradox in Section \ref{sec:analysis}. Related works and conclusion closes the paper.

\section{Differentiable Reasoning}
\label{diff_reasoning}


\subsection{Basics and Notation} 
\label{sec:basics}

We assume a knowledge base $\corpus$ is given in a relational logic language, where a formula $\varphi \in \corpus$ is built from predicate symbols $\mathsf{P} \in \mathcal{P}$, a finite set $\mathcal{D}$  of objects (also called constants) $o\in\mathbb{R}^m$  with $m \in \mathbb{Z}^{+}$, and variables $x \in \mathcal{V}$, in the usual way (see \cite{van2004logic}). We also assume that every $\varphi \in \corpus$ is in Skolem normal form. 
For a vector of objects and variables,  we use boldfaced $\mathbf{o}$ and $\mathbf{x}$, respectively. A ground atom is a formula with no logical connective and no variables, e.g.,  $\textsf{partOf}(cushion, chair)$ where $\textsf{partOf} \in \mathcal{P}$ and $cushion, chair \in \mathcal{D}$. Given a subset $D_i\subseteq \mathcal{D}$, a \textit{Herbrand base} $\bA_i$ corresponding to  $D_i$ is the set of all ground atoms generated from $D_i$ and  $\mathcal{P}$. A \textit{world} (often called a \textit{Herbrand interpretation}) $\mathbf{w}_i$ for $D_i$ assigns a binary truth value to each ground atom $\mathsf{P}(\mathbf{o})\in\mathbf{A}_i$ i.e., $\mathbf{w}_i(\mathsf{P}(\mathbf{o})) \in \{0, 1\}$. 

Each predicate $\mathsf{P}$ has a corresponding differentiable function $f_\mathsf{P}^{\btheta}(\mathbf{o}) \in \mathbb{R}^{\alpha(\mathsf{P})\times m} \rightarrow [0, 1]$ parameterized by $\btheta$ (a vector of reals) with $\alpha(\mathsf{P})$ being the arity of $\mathsf{P}$, which calculates the probability of $\mathsf{P}(\mathbf{o})$. This function could be, for instance, a neural network. 

Next, we define a Bernouilli distribution function over worlds as follows
\begin{equation}
\label{eq:pworld}
 	p(\mathbf{w}_i|\btheta, D_i) = \prod_{\mathsf{P}(\mathbf{o})\in\mathbf{A}_i}f_\mathsf{P}^{\btheta}(\mathbf{o})^{\mathbf{w}_i(\mathsf{P}(\mathbf{o}))}\cdot (1-f_\predP^{\btheta}(\mathbf{o}))^{1-\mathbf{w}_i(\mathsf{P}(\mathbf{o}))}
 \end{equation}
 where ${\mathbf{w}_i(\mathsf{P}(\mathbf{o}))}$ (similarly, ${1-\mathbf{w}_i(\mathsf{P}(\mathbf{o}))}$) refers to the exponent. Given some world $\mathbf{w_i}$, the \textit{valuation function} $v(\varphi, \mathbf{w_i})$ is 1 if $\varphi$ is true in that world, that is, $\bw_i \models \varphi$, and 0 otherwise.

Next, we explain the domain we use in this article.
We have a dataset $\mathcal{D}$ partitioned into two parts: a labeled dataset $\mathcal{D}_l = \langle\mathcal{O}_l, \mathcal{W}_l\rangle$ and an unlabeled dataset $\mathcal{D}_u = \langle \mathcal{O}_u , \emptyset \rangle$ 
where both $\mathcal{O}_l$ and $\mathcal{O}_u$ are sets of finite domains $D_i$, and $\mathcal{W}_l$ is a set containing the correct world $\mathbf{w}^{l*}_i$ for all pictures $i$. 

\begin{figure}
\includegraphics[width=\textwidth]{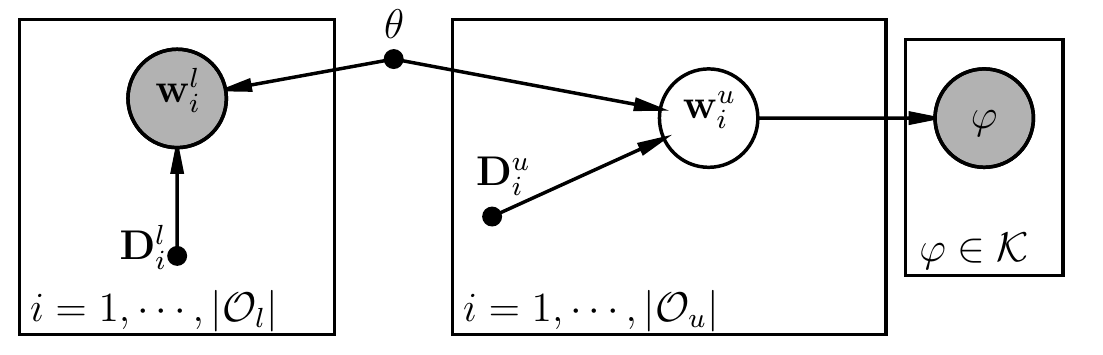}
\caption{The Bayesian network describing the joint probability $p(\mathcal{W}_l,\mathcal{W}_u,\mathcal{K}|\mathcal{O}_l,\mathcal{O}_u, \btheta)$. The left plate is the supervised classification likelihood and the right plates the unsupervised part in which we calculate the probability of the formulas $\mathcal{K}$. The parameters $\btheta$ are shared in both parts.}
\label{fig:pgm}
\end{figure}

In Figure \ref{fig:pgm} we illustrate the Bayesian network associated with this problem. The left plate denotes the usual supervised data likelihood $p(\mathcal{W}_l|\mathcal{O}_l, \btheta)$ and the right plates denote the probabilities of the truth values of the formulas $\varphi\in \mathcal{K}$ using $p(\mathcal{K}|\mathcal{O}_u, \btheta)$.


It is important to note that the true worlds $\mathbf{w}_i^{u*}$ of the unlabeled dataset are not known, that is, they are latent variables and they have to be marginalized over. The formulas in knowledge base  $\mathcal{K}$ are all assumed to be true. We can now obtain the optimization problem that we can solve using gradient descent as
\begin{align}
\btheta^* &= \argmax_{\btheta} p(\mathcal{W}_l|\mathcal{O}_l, \btheta)\cdot p(\mathcal{K}|\mathcal{O}_u, \btheta)\\
&= \argmax_{\btheta} \prod_{i=1}^{|\mathcal{O}_l|} p(\mathbf{w}^{l*}_i|D_i^l, \btheta) \cdot\prod_{i=1}^{|\mathcal{O}_u|}\sum_{\mathbf{w}_i^u}p(\mathbf{w}_i^u|D_i^u, \btheta)\cdot \prod_{\varphi\in \mathcal{K}}v(\varphi, \mathbf{w}^u_i)\\\begin{split}
\label{eq:optim}
&=\argmin_{\btheta} -\sum_{i=1}^{|\mathcal{O}_l|} \log p(\mathbf{w}^{l*}_i|D_i^l, \btheta)  \\ &  \qquad\qquad\qquad\qquad -\sum_{i=1}^{|\mathcal{O}_u|}\log\left(\sum_{\mathbf{w}_i^u}p(\mathbf{w}_i^u|D_i^u, \btheta) \cdot \prod_{\varphi\in \mathcal{K}}v(\varphi, \mathbf{w}^u_i)\right)
\end{split}
\end{align}
where in the last step we take the $\log$ and minimize with respect to the negative value. The optimization problem in Equation \ref{eq:optim} consists of two terms. The first is the cross-entropy loss for supervised labeled data. The second can be understood as follows: A world entails a (full) knowledge base (i.e., $\mathbf{w}\models \mathcal{K}$) if $\mathbf{w}\models \varphi$ holds for all $\varphi \in \mathcal{K}$ (that is, the product of their valuations is 1). For each domain $D_i$, we then find the sum of the probabilities of worlds that entail the knowledge base. This is an example of what we call the \textit{differentiable reasoning} loss. The general differentiable reasoning objective is given as 
\begin{equation}
\label{eq:diff_reason}
    \btheta^* = \argmin _{\btheta} -\sum_{i=1}^{|\mathcal{O}_l|} \log p(\mathbf{w}^{l*}_i|D_i^l, \btheta) + \mathcal{L}_{DR}(\btheta; \mathcal{K}, \mathcal{O}_u).
\end{equation}
\subsection{Differentiable Reasoning Using Product Real Logic}
\label{sec:prl}
The marginalization over all possible worlds $\mathbf{w}_i^u$ requires $2^{|\mathbf{A}_i|}$ combinations, so it is exponential in the size of the Herbrand base. Therefore, the problem of finding the sum of the probabilities $p(\mathbf{w}_i|\btheta)$ for all worlds $\mathbf{w}_i$ that entail the knowledge base $\mathcal{K}$ is \#P-complete \cite{roth1996hardness}
Instead, we shall perform a much simpler computation defined over logical formulas and the parameters $\btheta$ as follows:

\begin{align}
\mathcal{L}_{DR}(\btheta; \mathcal{K}, \mathcal{O}_u) &= \sum_{\varphi\in \mathcal{K}} \mathcal{L}(\btheta;\varphi,  \mathcal{O}_u)\\
\label{eq:forall}
\mathcal{L}(\btheta;\forall\bx \phi, \mathcal{O}_u)&=-\sum_{D\in \mathcal{O}_u, \bo\in D} \log \hat{p}(\phi|\bx=\bo, \btheta)\\
	\hat{p}(\textsf{P}(x_1, ..., x_{\alpha(\textsf{P})})|\bx = \mathbf{o}, \btheta) &= f_{\textsf{P}}^{\btheta}(o_1, ..., o_{\alpha(\textsf{P})})\\
    \hat{p}(\neg \phi|\bx = \mathbf{o}, \btheta) &= 1-\hat{p}(\phi|\bx = \mathbf{o}, \btheta)\\
	\hat{p}(\phi\wedge\psi|\bx = \mathbf{o}, \btheta) &= \hat{p}(\phi|\bx = \mathbf{o}, \btheta) \cdot \hat{p}(\psi|\bx = \mathbf{o}, \btheta) \\
    \hat{p}(\phi\vee\psi|\bx = \mathbf{o}, \btheta) &= \hat{p}(\neg(\neg\phi\wedge\neg\psi)|\bx=\bo, \btheta)\\
    \hat{p}(\phi\rightarrow\psi|\bx = \mathbf{o}, \btheta) &= \hat{p}(\neg\phi\vee \psi|\bx = \mathbf{o}, \btheta)
\end{align}
where $\alpha:\mathcal{P} \rightarrow \mathbb{Z}^+$ is the arity function for each predicate symbol, and $\phi$ and $\psi$ are subformulas of $\varphi$.  
$\prl$ computes the fuzzy degree of truth of some formula $\varphi$ using the product norm and the Reichenbach implication \cite{bergmann2008introduction}, which makes our approach a special case of Real Logic \cite{serafini2016logic} that we call \textit{Product Real Logic}. The $\forall$ quantifier is interpreted in Equation $\ref{eq:forall}$ by going through all instantiations, which in this case is all $n$-tuples in the domain $D_i$, and also looping over all domains $D_i$ in the set of domains (i.e., pictures) $\mathcal{O}_i$. 

\begin{wrapfigure}{R}{0.4\textwidth}
    \centering
    \includegraphics[width=0.36\textwidth]{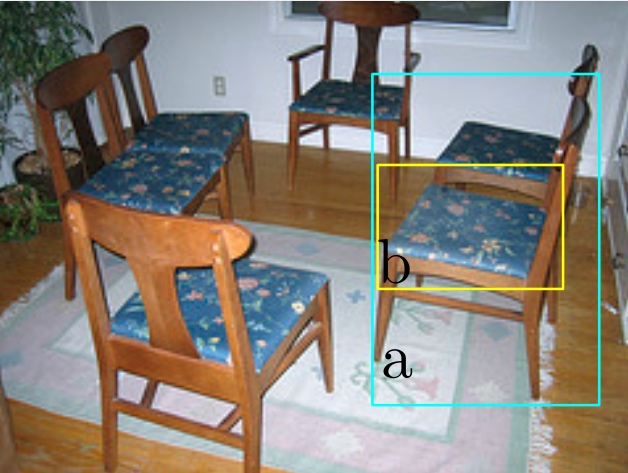}
    \caption{We can deduce that $b$ is a cushion if we are confident about the truth value of $\textsf{chair}(a)$ and $\textsf{partOf}(b, a)$ using the formula $\forall x, y\ \textsf{chair}(x)\wedge \textsf{partOf}(y, x)\rightarrow \textsf{cushion}(y)$.}
    \label{fig:example_dr}
\end{wrapfigure}

\begin{exmp}
\label{exmp:rl}
The loss term associated with the formula $\varphi=\forall x, y\ \textsf{chair}(x)\wedge \textsf{partOf}(y, x)\rightarrow \textsf{cushion}(y)\vee\pred{armRest}(y)$ is computed as follows:
\begin{align*}
\loss(\btheta;\varphi, \mathcal{O}_u) = -\sum_{D\in \mathcal{O}, o_1, o_2\in D} &1- f_\textsf{chair}^{\btheta}(o_1)\cdot f^{\btheta}_{\textsf{partOf}}(o_2, o_1)\cdot\\ (&1-f^{\btheta}_{\textsf{cushion}}(o_2))\cdot(1-f^{\btheta}_{\textsf{armRest}}(o_2))
\end{align*}

Say $\mathcal{O}_u$ contains the picture in Figure \ref{fig:example_dr} whose domain is $\{a, b\}$ and the model predicts the following distribution over worlds:
\begin{align*}
\fp{chair}(a)&=0.9  & \fp{chair}(b)&=0.4\\
\fp{cushion}(a)&=0.05 & \fp{cushion}(b)&=0.5\\
\fp{armRest}(a)&=0.05 & \fp{armRest}(b)&=0.1\\
\fp{partOf}(a,a) &= 0.001 & \fp{partOf}(b,b) &= 0.001\\
\fp{partOf}(a,b) &= 0.01 & \fp{partOf}(b,a) &= 0.95
\end{align*}

The model returns high values for $\fp{chair}(a)$ and $\fp{partOf}(b, a)$ but it is not confident of $\fp{cushion}(b)$, even though it is clearly higher than $\fp{armRest}(b)$. We can decrease $\loss(\btheta;\varphi, \mathcal{O}_u) = 0.612$ simply by increasing $\fp{cushion}(b)$, since $f^{\btheta}_{\textsf{cushion}}$ is a differentiable function with respect to $\btheta$.

This example shows that we can find a new instance of the \textsf{cushion} predicate using reasoning on an unlabeled dataset. This process uses both statistical reasoning and symbolic rules. As more data improves generalization, those additional examples could help reducing the sparsity of the SII problem. Furthermore, \cite{donadello2017logic} showed that it is also possible to correct wrong labels due to noisy data when these do not satisfy the formulas. 
\end{exmp}

\begin{figure}
    \centering
    \includegraphics[width=0.7\textwidth]{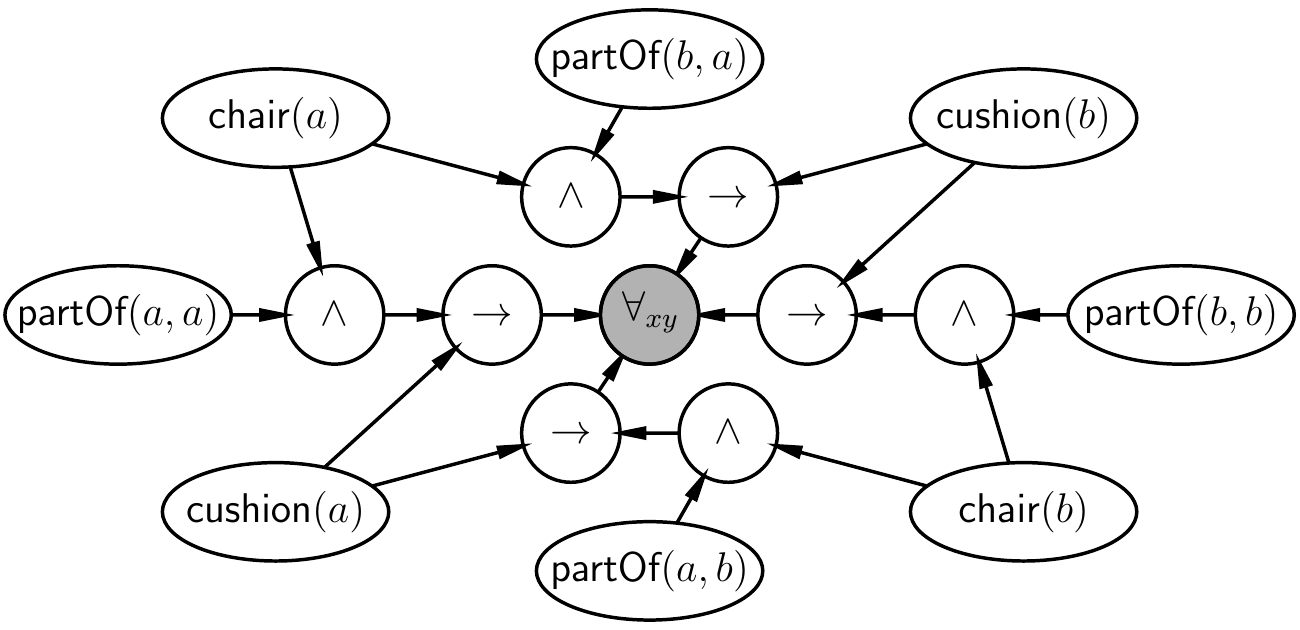}
    \caption{The Bayesian network associated with grounding of the formula $\forall x, y\ \textsf{chair}(x)\wedge \textsf{partOf}(y, x)\rightarrow \textsf{cushion}(y)$ on the domain from Figure \ref{fig:example_dr}. We treat connectives and quantifiers as  binary random variables (which correspond to subformulas through their parents) of which the conditional probabilities are computed using truth tables.}
    \label{fig:example_pgm}
\end{figure}

Figure \ref{fig:example_pgm} shows the Bayesian Network for this formula on the picture from Figure \ref{fig:example_dr}, illustrating the computation path. We treat each subformula as a binary random variable of which the conditional probabilities are given by truth tables. Because the graph is not acyclic, we can use loopy belief propagation which is empirically shown to often be a good approximation of the correct probability \cite{Murphy2013}. In fact, Product Real Logic can be seen as performing a single iteration of belief propagation. However, this can be problematic. For example, the degree of truth of the ground formula $\textsf{chair}(o)\wedge\textsf{chair}(o)$ would be computed using $f_\textsf{chair}^{\btheta}(o)^2$ instead of the probability of this statement, $f_\textsf{chair}^{\btheta}(o)$ \cite{rocktaschel2015injecting}. We show in Appendix \ref{appendix:deriv} that Product Real Logic computes the correct probability $p(\corpus|\mathcal{O}_u, \btheta)$ for a corpus $\corpus$ under the strong assumption that, after grounding, each ground atom is used at most once.

An interesting and useful property of our approach is that it can perform multi-hop reasoning in an iterative, yet extremely noisy, manner. In one iteration it might, for instance, increase $f_\textsf{cushion}^{\btheta}(o)$. And  since $f_\textsf{cushion}^{\btheta}(o)$ will return higher values in future iterations, it can be used to prove that the probability of other ground atoms that occur in formulas with $\textsf{cushion}(o)$ should also be increased or decreased.

A convenient property of the SII task is that we consider just binary relations between objects appearing on the same pictures. The Herbrand base then contains $O(|\mathcal{P}|\cdot |D_i|^2)$ ground atoms, which is feasible as there are often not more than a few dozen objects on an image. This property also holds in natural language to some degree in the following way: only the words appearing in the same paragraph can be related. This is in contrast to the knowledge base completion task where we have a single graph with many objects and predicates \cite{socher2013reasoning}.

\subsection{Implementation}
We optimize the negative logarithm of the likelihood function given in Equation \ref{eq:optim}. In particular, we use minibatch gradient descent to decrease the computation time 
both for the supervised part of the loss and the unsupervised part. We turn the unsupervised loss into minibatch gradient descent by approximating the computation of the $\forall$ quantifier: instead of summing over all $n$-tuples and all domains, we randomly sample from these $n$-tuples independently from the domain it belongs to. 
\subsection{The Material Implication}
\label{sec:raven}
To provide a better understanding of the inner machinery of our approach, we will elaborate on some interesting partial derivatives. Say, we have a formula $\varphi$ of the form $\forall {\bx}\phi(\bx) \rightarrow \psi(\bx)$, where $\phi(\bx)$ is the antecedent and $\psi(\bx)$ the consequent of $\varphi$. First, we write out the partial derivative of $\loss(\btheta; \varphi, \mathcal{O}_u)$ with respect to the consequent, where we make use of the chain rule:
\begin{align}
    \dmp(\bo) &:= \frac{\partial \log \hat{p}(\varphi| \mathcal{O}_u, \btheta)}{\partial \ppsi} = \frac{\partial \sum_{\bo \in D, D\in \mathcal{O}_u} \log \prl(\phi\rightarrow\psi|\bo, \btheta)}{\partial \ppsi}\\ 
    &= \frac{\partial\sum_{\bo \in D, D\in \mathcal{O}_u} \log(1-\prl(\phi|\bo, \btheta)\cdot(1-\prl(\psi|\bo, \btheta)))}{\partial \ppsi} \\
    \label{eq:dmp}
    &=\frac{\hat{p}(\phi|\bo, \btheta)}{1-\prl(\phi|\bo, \btheta)\cdot(1-\prl(\psi|\bo, \btheta)))}=\frac{\hat{p}(\phi|\bo, \btheta)}{\hat{p}(\phi\rightarrow \psi|\bo, \btheta)}
\end{align}

$\dmp(\bo)$ mirrors the application of the Modus Ponens (MP) rule using the implication $\phi\rightarrow \psi$ for the assignment of $\bo$ to $\bx$. The MP rule says that if $\phi$ is true and $\phi\rightarrow\psi$, then $\psi$ should also be true. Similarly, if $\phi(\bo)$ is likely and $\phi\rightarrow \psi$, then $\psi(\bo)$ should also be likely. Indeed, notice that $\dmp(\bo)$ grows with $\pphi$. Also,   $\dmp(\bo)$ is largest when $\pphi$ is high and $\ppsi$ is low as it then approaches a singularity in the divisor. 
We next show the derivation with respect to the negated antecedent:

\begin{align}
\label{eq:dmt}
    \dmt(\bo) &:= \frac{\partial \log \hat{p}(\varphi| \mathcal{O}_u, \btheta)}{\partial \pnphi} =  \frac{\hat{p}(\neg\psi|\bo, \btheta)}{\hat{p}(\phi\rightarrow \psi|\bo, \btheta)}
\end{align}

Similarly, it mirrors the application of the Modus Tollens (MT) rule which says that if $\psi$ is false and $\phi\rightarrow\psi$, then $\phi$ should also be false. Again, realize that $\dmt(\bo)$ grows with $\ppsi$.

It is easy to see that $\dmp(\bo)>\dmt(\bo)$ whenever $\pphi > \pnpsi$. Furthermore, the global minimum of $\loss(\btheta;\varphi, \mathcal{O}_u)$ is some parameter value $\btheta^*$ so that $\prl(\phi|\mathcal{O}_u, \btheta^*)=0$ and $\prl(\psi|\mathcal{O}_u, \btheta^*)=1$ for all $\bo$, which corresponds to the material implication.

Next, we show how these quantities are used in the updating of the parameters $\btheta$ using backpropagation and act as mixing components on the gradient updates:
\begin{align}
\label{eq:deriv}
    \frac{\log \hat{p}(\varphi| \mathcal{O}_u, \btheta)}{\partial \btheta} = \sum_{\bo \in D, D\in \mathcal{O}_u} d^{\mathrm{MP}}_\varphi(\bo) \cdot \frac{\partial \ppsi}{\partial \btheta} + d^{\mathrm{MT}}_\varphi(\bo) \cdot \frac{\partial \pnphi}{\partial \btheta}
\end{align}
\subsection{The Raven Paradox}
In our experiments, we have found that this approach is very sensitive to the \emph{raven paradox} \cite{hempel1945studies}. It is stated as follows: Assuming that observing an example of a statement is evidence for that statement (i.e., the degree of belief in that statement increases), and that evidence for a sentence also is evidence for all the other logically equivalent sentences, then our belief in ``ravens are black'' increases when we observe non-black non-raven, by the contrapositive ``non-ravens are non-black''. Equation \ref{eq:deriv} shows however that the gradient is equally determined by positive evidence (observing black ravens) as by contrapositive evidence (observing non-black non-ravens). Because in the real world there are far more ravens than non-black objects, optimizing $\prl(\forall o\ \textsf{raven}(o) \rightarrow \textsf{black}(o)|\mathcal{O}_u, \btheta)$ amounts to recognizing that something is not a raven when it is not black. However, Machine Learning models tend to be biased when the class distribution is unbalanced during training \cite{Weiss2001}.

\begin{figure}%
    \centering
    \subfloat{{\includegraphics[width=6cm]{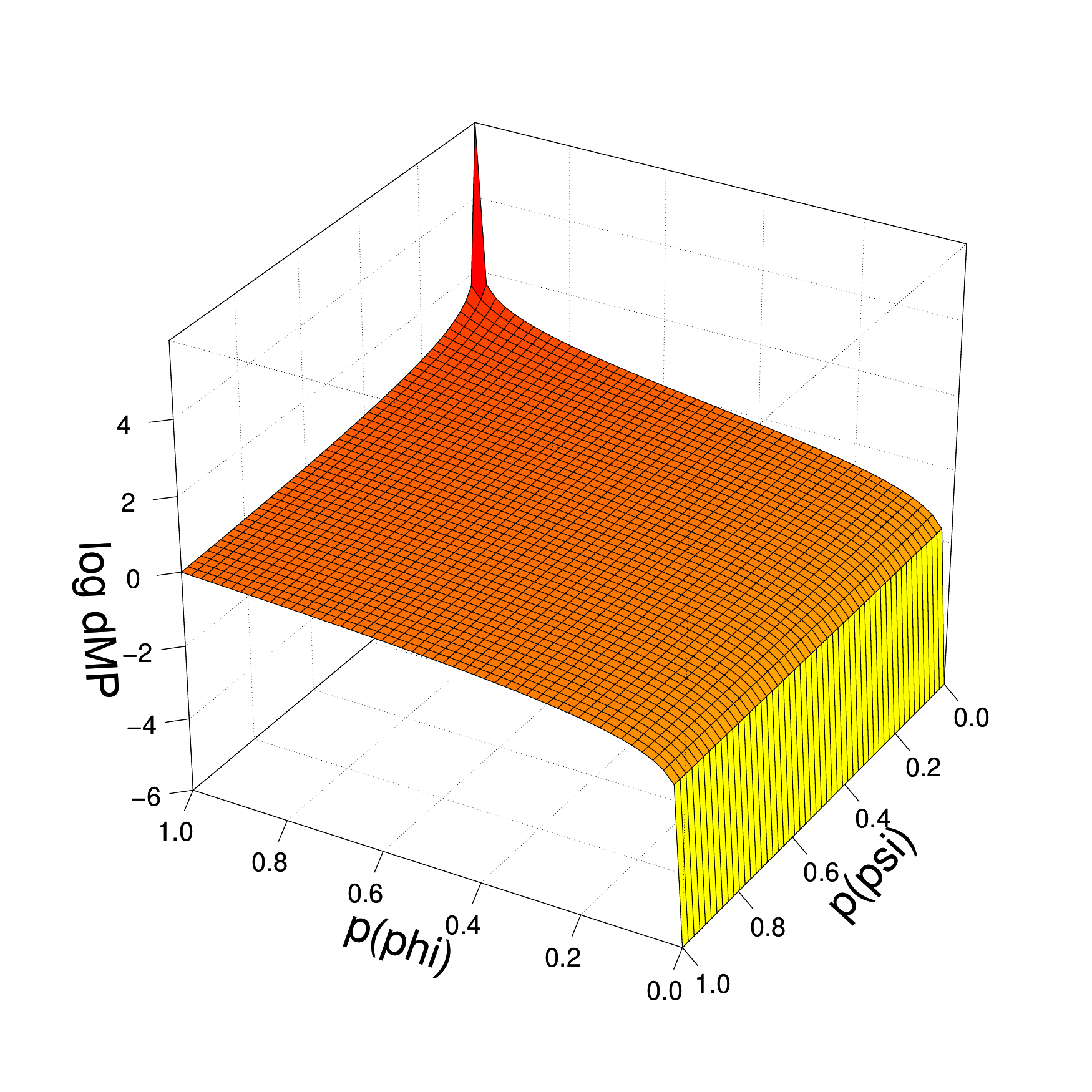} }}%
    \subfloat{{\includegraphics[width=6cm]{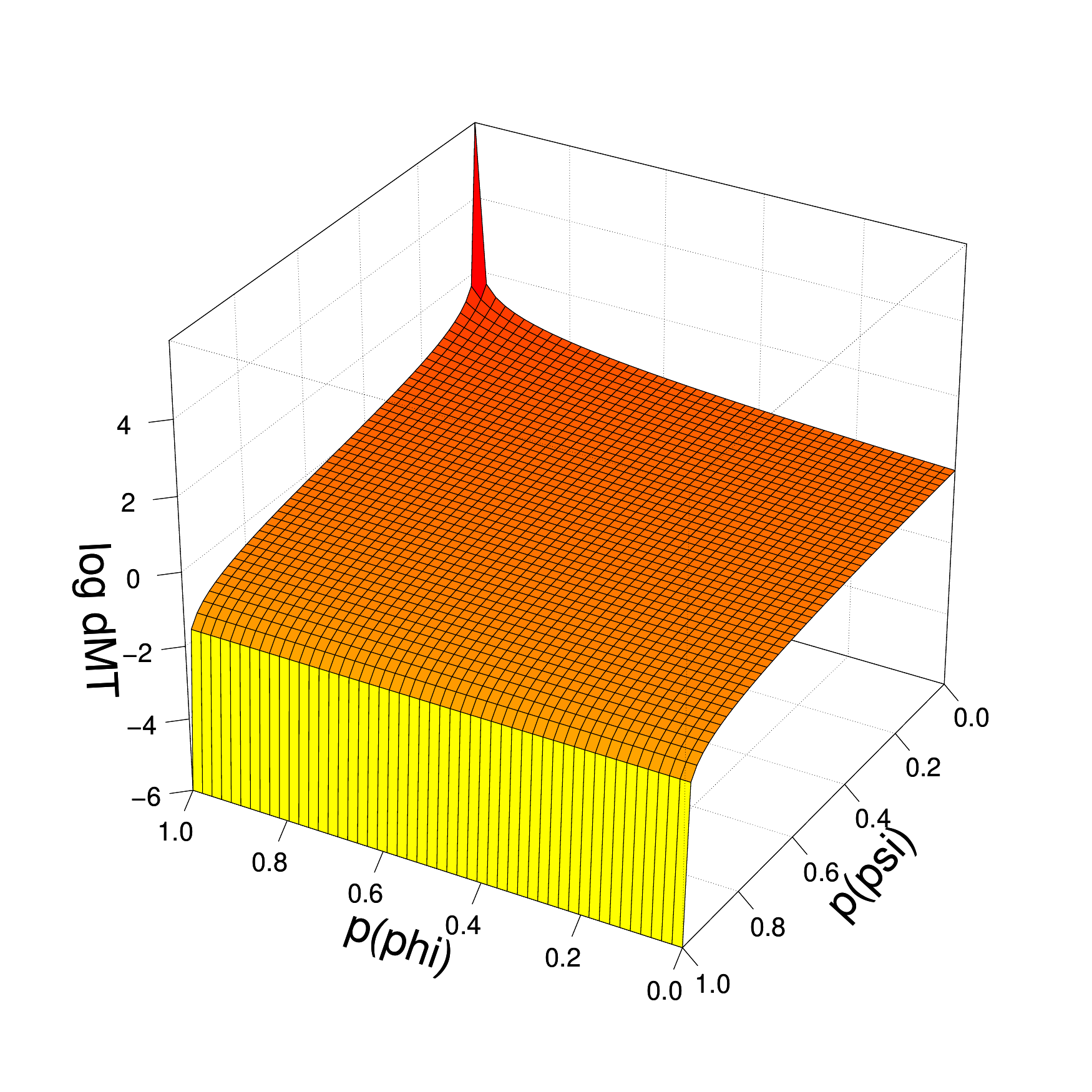} }}%
    \caption{Plots of $\dmp(\bo)$ (Equation \ref{eq:dmp}) and $\dmt(\bo)$ (Equation \ref{eq:dmt}). Note that the y axis is using a log scale.}%
    \label{fig:dmpmt}%
\end{figure}

Figure \ref{fig:dmpmt} shows plots of $\dmp(\bo)$ and $\dmt(\bo)$ for different values of $\pphi$ and $\ppsi$. In practice, for many formulas of this form, the most common case will be that the model predicts $\neg\phi(\bo)\wedge\neg\psi(\bo)$. Then, $\dmp(\bo)$ approaches 0 and $\dmt(\bo)$ will be around 1. For instance, the average value of $\dmp(\bo)$ for the problem in Example \ref{exmp:rl} is $0.214$, while the average value of $\dmt(\bo)$ is $0.458$.

We analyze a naive way of dealing with this phenomenon. We normalize the contribution to the total gradient of MP and MT reasoning by replacing the loss function $\mathcal{L}$ of rules of the form $\forall {\bx}\phi(\bx) \rightarrow \psi(\bx)$ as follows:
\begin{equation}
\begin{split}
\label{eq:normalize}
    \mathcal{L}(\btheta; \corpus, \mathcal{O}_u) = -\sum_{\varphi\in\corpus}\sum_{\bo \in D, D\in \mathcal{O}_u} &\frac{\mu \cdot d^{\mathrm{MP}}_\varphi(\bo)}{\sum_{\bo' \in D, D\in \mathcal{O}_u}d^{\mathrm{MP}}_{\varphi}(\bo')} \cdot \ppsi \\ &+\frac{(1-\mu) \cdot d^{\mathrm{MT}}_\varphi(\bo)}{\sum_{\bo' \in D, D\in \mathcal{O}_u}d^{\mathrm{MT}}_{\varphi}(\bo')} \cdot \pnphi
\end{split}
\end{equation}

where $\mu$ is a hyperparameter that assigns the relative importance of Modus Ponens with respect to Modus Tollens updates. We are then able to control how much either contributes to the training process. We experiment with different values of $\mu$ and report our findings in the next section.

\section{Experiments}
\label{sec:experiments}
We carried out simple experiments on the \textsc{PASCAL-Part} dataset \cite{chen2014detect} in which the task is to predict the type of the object in a bounding box and the $\textsf{partOf}$ relation which expresses that some bounding box is a part of another. For example, a tail can be a part of a cat. Like in \cite{donadello2017logic}, the output softmax layer over the 64 object classes of a Fast R-CNN \cite{girshick2015fast} detector is used for the bounding box features. Note that this makes the problem of recognizing types very easy as the features correlate strongly with the true output types. Therefore, to get a more realistic estimate, we randomly split the dataset into only 7 labeled pictures for $\mathcal{D}_l$ and 2128 unlabeled pictures for $\mathcal{D}_u$. Additionally, we only consider 11 (related) types out of 64 due to computational constraints. As there is a large amount of variance associated with randomly splitting in this way, we do all our experiments on 20 random splits of the dataset. The results are evaluated on a held-out validation set of 200 images. We compare the accuracy of prediction of the type of the bounding box and the AUC (area under curve) for the $\textsf{partOf}$ relationship.

We model $f_{\textsf{type}_i}^{\btheta}(o)$ using a single Logic Tensor Network (LTN) layer \cite{donadello2017logic} of width 10 followed by a softmax output layer to ensure mutual exclusivity of types. The term $f_{\textsf{partOf}}^{\btheta}(o_1, o_2)$ is modeled using an LTN layer of width 2 and a sigmoid output layer. The loss function is then optimized using RMSProp over 6000 iterations. We use the same relational background knowledge as \cite{donadello2017logic} which are rules like the following:


\begin{table}[h]
\centering
\label{Rules}
\begin{tabular}{|l|}
\hline
$\forall x, y\ \pred{chair}(x) \wedge \pred{partOf}(y, x) \rightarrow \pred{cushion}(y)\vee\pred{armRest}(y) $\\
$\forall x, y\ \pred{cushion}(x) \wedge \pred{partOf}(x, y) \rightarrow \pred{chair}(y) \vee \pred{bench}(y) $\\
$\forall x\ \neg \textsf{partOf}(x, x)$ \\
$\forall x, y\ \textsf{partOf}(x, y) \rightarrow \neg \textsf{partOf}(y, x)$ \\
    \hline
\end{tabular}
\end{table}
\begin{wraptable}{r}{0.52\textwidth}
\centering
\begin{tabular}{|l|l|}
\hline
                    & Precision types  \\ \hline
Supervised            & $0.440 \pm 0.0013$  \\ \hline
Unnormalized  & $0.455 \pm 0.0014$    \\ \hline
Normalized $\mu=0$ & $0.454 \pm 0.0015$ \\ \hline
Normalized $\mu=0.1$ & $0.505 \pm 0.0014$ \\ \hline
Normalized $\mu=0.25$ & $\textbf{0.517} \pm 0.0013$ \\ \hline
Normalized $\mu=0.5$ & $0.510 \pm 0.0013$ \\ \hline
Normalized $\mu=0.75$ & $0.496 \pm 0.0012$ \\ \hline
Normalized $\mu=1$ & $0.435 \pm 0.0015$  \\ \hline
\end{tabular}
\caption{Results of the experiments. 20 runs using random splits of the data are averaged alongside 95\% confidence intervals. All results are significant.}
\label{table:results}
\end{wraptable}
We compare three methods. In the first one we train without any rules, which forms the \emph{supervised} baseline. In the second, \emph{unnormalized}, we add the rules to the unlabeled data. This does not use any technique for dealing with the raven paradox. In the last one called \emph{normalized}, we normalize MP and MT reasoning using Equation \ref{eq:normalize} for several different values of $\mu$. The results in Table \ref{table:results} are statistically significant when using a paired t-test.
\WFclear
\section{Analysis}
\label{sec:analysis}
Our experiments show that we can significantly improve on the classification of the types of objects for this problem. The normalized method in particular outperforms the unnormalized method, suggesting that explicitly dealing with the raven paradox is essential in this problem. 

\subsection{Gradient Updates}
\begin{wrapfigure}{r}{0.5\textwidth}
        \includegraphics[width=.5\textwidth]{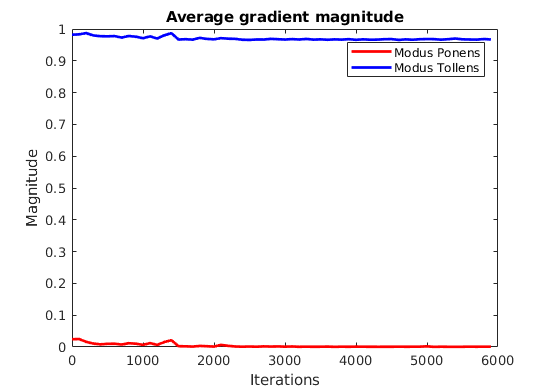}
        \caption{The average magnitude of Modus Ponens and Modus Tollens gradients.}
    \label{fig:grad_magn}
\end{wrapfigure}

We analyze how the different methods handle the implication using the quantities $\dmp$ and $\dmt$ defined in Section \ref{sec:raven}. Figure \ref{fig:grad_magn} shows the average magnitude of $\dmp$ and $\dmt$ in the unnormalized model, which is computed by averaging over all training examples and formulas. This shows that the average MT gradient update is, in this problem, around 100 times larger than the average MP gradient update, i.e., it uses far more contrapositive reasoning. The unnormalized method acts very similar to the normalized one with $\mu\approx 0.01$.

Next, we will analyze how accurate our approach is at reasoning by comparing its 'decisions' to what should have been the correct 'decision'. We sample 2000 pairs of bounding boxes from the \textsc{PASCAL-Part} test set $\langle\mathcal{O}_t, \mathcal{W}_t\rangle$. We consider a pair of bounding boxes $\bo$ from an image $i$ in the test set $\mathcal{O}_t$. $d^{\mathrm{MP}}_\varphi(\bo)$ is a \textit{correctly reasoned gradient} if both $\phi(\bo)$ and $\psi(\bo)$ are true in $\bw_i^t$. Likewise, $d^{\mathrm{MT}}_\varphi(\bo)$ is a correctly reasoned gradient if $\neg\psi(\bo)$ and $\neg\phi(\bo)$ are true in $\bw_i^t$. Furthermore, we say that $d^{\mathrm{MP}}_\varphi(\bo)$ is a \textit{correctly updated gradient} if at least $\psi(\bo)$ is true in $\bw_i^t$, and that $d^{\mathrm{MT}}_\varphi(\bo)$ is correctly updated when $\neg \phi(\bo)$ is true in $\bw_i^t$. Then the \textit{correctly reasoned ratios} are computed using
\begin{align}
\label{eq:cr_mp}
\mathrm{cr}^{\mathrm{MP}} &= \frac{\sum_{\varphi \in \corpus}\sum_{\bo\in D_i, D_i\in \mathcal{O}_t} v(\phi, \bw_i^t)\cdot v(\psi, \bw_i^t)\cdot \dmp(\bo)}{\sum_{\varphi \in \mathcal{K}}\sum_{\bo'\in D_i, D_i\in \mathcal{O}_t}\dmp(\bo')}\\
\label{eq:cr_mt}
\mathrm{cr}^{\mathrm{MT}} &= \frac{\sum_{\varphi \in \corpus}\sum_{\bo\in D_i, D_i\in \mathcal{O}_t} v(\neg\phi, \bw_i^t)\cdot v(\neg\psi, \bw_i^t)\cdot d^{\mathrm{MT}}_\varphi(\bo)}{\sum_{\varphi \in \corpus}\sum_{\bo'\in D_i, D_i\in \mathcal{O}_t}d^{\mathrm{MT}}_\varphi(\bo')}.
\end{align}

The definition of the \textit{correctly updated ratios} ($\mathrm{cu}^{\mathrm{MP}}$ and $\mathrm{cu}^{\mathrm{MT}}$) are nearly the same. $\mathrm{cu}^{\mathrm{MP}}$ is found by removing the $v(\phi, \bw_i^t)$ term from Equation \ref{eq:cr_mp}, and $\mathrm{cu}^{\mathrm{MT}}$ by removing the $v(\neg \psi, \bw_i^t)$ term from Equation \ref{eq:cr_mt}. 

\begin{figure}
\centering
    \includegraphics[width=1\textwidth]{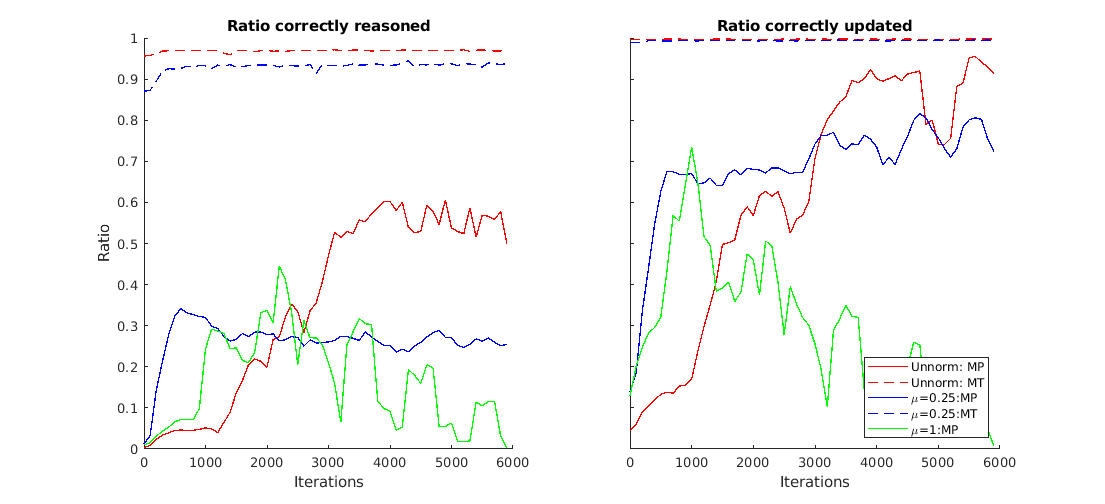}
    \caption{The left plot shows $\mathrm{cr}^{\mathrm{MP}}$ and $\mathrm{cr}^{\mathrm{MT}}$ and the right plot $\mathrm{cu}^{\mathrm{MP}}$ and $\mathrm{cu}^{\mathrm{MT}}$ for the Unnormalized method (denoted as Unnorm) and the Normalized methods with $\mu=0.25$ and $\mu=1$.}
    \label{fig:good_updates}
\end{figure}

Figure \ref{fig:good_updates} shows the value of these ratios during training. The dotted lines that represent MT reasoning shows a convenient property, namely that is nearly always correct because of the large class imbalance. This could be the reason there is a significant benefit to adding contrapositive reasoning. Both normalized and unnormalized at $\mu=1$ seems to get 'better' at reasoning during training, as the correctly updated ratios go up. After training for some time, the unnormalized method seems to be best at reasoning correctly for both MP and MT. Another interesting observation is the difference between $\crmp$ and $\cump$. At many points, about half of the gradient magnitude correctly increases $\hat{p}(\psi|\bo,\btheta)$ because the model predicts a high value for $\prl(\phi|\bo, \btheta)$, even though $\phi(\bo)$ is not actually in the test labels. It is interesting to see that, this kind of faulty reasoning which does lead to the right conclusion is actually beneficial for training.

Furthermore, disabling MT completely by setting $\mu$ to 1 seems to destabilize the reasoning. This is also reflected in the validation accuracy that seems to decline when $\mathrm{cu}^{\mathrm{MP}}$ declines. This suggests that contrapositive reasoning is required to increase the amount of correct gradient updates.

\section{Related work}
\subsection{Injecting Logic into Parameterized Models}
Our work follows the recent works on Real Logic \cite{serafini2016logic,donadello2017logic}, and the method we use is a special case of Real Logic with some additional changes. A particular difference is that the logic we employ has no function symbols, which was due to simplicity purposes.   Injecting background knowledge into vector embeddings of entities and relations has been studied in \cite{Demeester2016,demeester2016regularizing,rocktaschel2017combining,rocktaschel2017end}. In particular, \cite{rocktaschel2015injecting} has some similarities with Real Logic and our method. However, this method is developed for regularizing vector embeddings instead of any parameterized model.  In this sense, it can also be seen as a special case of Real Logic. Semantic Loss \cite{Xu2017} is a very similar semi-supervised learning method. This loss is essentially Equation \ref{eq:optim}, which makes it more accurate than Product Real Logic, but also exponential in runtime. To deal with this, they compile SDD's \cite{Darwiche2011} to make the computation tractable. A recent direction is DeepProbLog \cite{manhaeve2018deepproblog}, a probabilistic version of Prolog with neural predicates that also uses SDD's. \cite{hu2016harnessing} also injects rules into a general model with a framework that transfers the logic rules using a so-called teacher network. This model is significantly different from the aforementioned ones, as it does not add a loss for each rule. 

\subsection{Semi-Supervised Learning}
There is a large body of literature on semi-supervised methods \cite{oliver2018, chapelle2009semi}. In particular, recent research on graph-based semi-supervised learning \cite{kipf2016semi,yang2016revisiting, zhu2003semi} relates unlabeled and labeled data through a graph structure. However, they do not use logically structured background knowledge. It is generally used for entity classification, although in \cite{schlichtkrull2017modeling} it is also used on link prediction. \cite{lee2013pseudo} introduced the surprisingly effective method Pseudo-Label that first trains a model using the labeled dataset, then labels the unlabeled dataset using this model and continues training on this newly labeled dataset. Our approach has a similar intuition in that we use the current model to get an estimation about the correct labels of the labeled dataset, and then use those labels to predict remaining labels, but the difference is that we use background knowledge to choose these labels.

\section{Conclusion and Future Work}
We proposed a novel semi-supervised learning technique and showed that it is possible to find labels for samples in an unlabeled dataset by evaluating them on relational background knowledge. Since implication is at the core of logical reasoning, we analyzed this by inspecting the gradients with respect to the antecedent and the consequent. Surprisingly, we discovered a strong imbalance between the contributions of updates from MP and MT in the induction process. It turned out that our approach is highly sensitive to the Raven paradox \cite{hempel1945studies} requiring us to handle positive and contrapositive reasoning separately. Normalizing these different types of reasoning  yields the largest improvements to the supervised baseline. Since it is quite general, we suspect that issues with this imbalance could occur in many systems that perform inductive reasoning.

We would like to investigate this phenomenon with different background knowledge and different datasets such as VisualGenome and ImageNet. In particular, we are interested in other approaches for modelling the implication like different Fuzzy Implications \cite{Jayaram2008} or by taking inspiration from Bayesian treatments of the Raven paradox \cite{Vranas2004}. Furthermore, it could be applied to natural language understanding tasks like semantic parsing.

\medskip

\small

\bibliographystyle{plain}
\bibliography{references.bib}

\begin{appendices}
\section{Conditional Optimality of Product Real Logic}
\label{appendix:deriv}

Considering only a single domain $D$ of objects $x\in \mathbb{R}^D$, we have the Herbrand base $\bA$. Let $\varphi\in \corpus$ be a set of function-free FOL formulas in Skolem-normal form. Furthermore, let $\mathcal{P}=\{\predP_1, ..., \predP_K\}$ be a set of predicates which for ease of notation and without loss of generality we assume to all have the arity $\alpha$. 

Each ground atom $\predP(\bo)\sim\mathrm{Bern}(f_\predP^{\btheta}(\bo))$ is a binary random variable that denotes the binary truth value. It is distributed by a Bernoulli distribution with mean $f^{\btheta}_\predP\in \mathbb{R}^{\alpha\times D}\rightarrow [0, 1]$.

For each formula $\varphi$, we have the set of ground atoms $\bA_\varphi\subseteq \bA$ appearing in the instantiations of $\varphi$. Likewise, the assignment of truth values of $\bA_\varphi$ is $\bw_\varphi$, which is a subset of the world $\bw$. 
We can now express the joint probability, using Equation \ref{eq:pworld} and the valuation function defined in Section \ref{sec:basics}:
\begin{align}
p(\corpus, \bw|D, \btheta)&=p(\bw|\btheta)\cdot\prod_{\phi\in\corpus} v(\phi, \bw_\varphi)
\end{align}
We will first show that Product Real Logic is equal to this probability with two strong assumptions. The first is that the sets of ground atoms $\bA_\varphi$ are disjoint for all formulas in the corpus, i.e. if
\begin{equation}
\label{eq:disjoint}
    \bigcup_{\varphi\in\corpus}\bA_\varphi = \emptyset
\end{equation}

The second is that the set of ground atoms used in two children (a direct subformula) of some subformula of a formula in $\corpus$ are disjoint. If $pa(\phi)$ returns the parent of $\phi$ and $r(\phi)$ returns the root of $\phi$ (the formula highest up the tree), then
\begin{equation}
\label{eq:disjoint_sub}
    \bA_{\phi} \cup \bA_{\psi} = \emptyset, \forall \{\phi, \psi |pa(\phi) = pa(\psi) \wedge r(\phi)\in\corpus\}
\end{equation}

First, we marginalize over the different possible worlds:
\begin{flalign}
    \label{eq:deriv1}
    p(\corpus|D, \btheta) &= \sum_\bw p(\bw|\btheta) \cdot \prod_{\varphi\in\corpus} v(\phi, \bw_\varphi) \\
    &= \sum_{\bw_{\varphi_1}} p(\bw_{\varphi_1}|\btheta) \cdot \left(... \cdot \sum_{\bw_{\varphi_{|\corpus|}}} p(\bw_{\varphi_{|\corpus|}}|\btheta) \cdot \prod_{\varphi\in\corpus} v(\phi, \bw_\varphi)\right) \\
    &= \sum_{\bw_{\varphi_1}} p(\bw_{\varphi_1}|\btheta)\cdot v(\varphi_1, \bw_{\varphi_1}) \cdot \left( ... \cdot \sum_{\bw_{\varphi_{|\corpus|}}} p(\bw_{\varphi_{|\corpus|}}|\btheta) \cdot v(\varphi_{|\corpus|}, \bw_{\varphi_{|\corpus|}}) \right)\\
    \label{eq:deriv2}
    &= \prod_{\varphi\in\corpus} \sum_{\bw_\varphi} p(\bw_\varphi|\btheta) \cdot  v(\phi, \bw_\varphi)
\end{flalign}
where we make use of Equation \ref{eq:disjoint} to join the summations, the independence of the probabilities of atoms from Equation \ref{eq:pworld} and marginalization of the atoms other than those in $\bA_\varphi$.

 We denote the set of instantiations of $\varphi$ by $S_\varphi$, and a particular instance by $s$. $\bA_s\subseteq\bA_\varphi$ then is the set of ground atoms in $s$ (and respectively for $\bw_s$. Next we show that $\sum_{\bw_\varphi} p(\bw_\varphi|\btheta) \cdot  v(\varphi, \bw_\varphi)=\prod_{s\in S_\varphi}\prl(\varphi|s, \btheta)$. As the formulas are in prenex normal form, $\varphi=\forall x_1, ..., x_\alpha \phi$. We find that, using Equation \ref{eq:disjoint_sub} and the same procedure as in Equations \ref{eq:deriv1}-\ref{eq:deriv2}
\begin{align}
    \sum_{\bw_\varphi} p(\bw_\varphi|\btheta) \cdot  v(\varphi, \bw_\varphi) &= \sum_{\bw_\varphi} p(\bw_\varphi|\btheta) \cdot \prod_{s\in S_\varphi} v(\phi, \bw_s)\\
    \label{eq:deriv3}
    &= \prod_{s\in S_\varphi} \sum_{\bw_s} p(\bw_s|\btheta) \cdot v(\phi, \bw_s).
\end{align}

Then, it suffices to show that $\sum_{\bw_s} p(\bw_s|\btheta) \cdot  v(\phi, \bw_s)=\prl(\phi|s, \btheta)$. This is done using recursion. For brevity, we will only proof it for the $\neg$ and $\wedge$ connectives, as we can proof the others using those.

Assume that $\phi=\predP(x_1, ..., x_n)$. Then if $w_s(\predP(x_1, ..., x_n))$ is the binary random variable of the ground atom $\predP(x_1, ..., x_n)$ under the instantiation $s$, 
\begin{align}
    &\sum_{\bw_s} p(\bw_s|\btheta) \cdot  v(\predP(x_1, ..., x_n), \bw_s) \\ 
    =& \sum_{\bw_s \backslash\{w_s(\predP(x_1, ..., x_n))\}} p(\bw_s \backslash\{w_s(\predP(x_1, ..., x_n))\}|\btheta) \cdot\\ & \sum_{w_s(\predP(x_1, ..., x_n))} p(w_s(\predP(x_1, ..., x_n))|\btheta)\cdot w_s(\predP(x_1, ..., x_n))\\
    =& p(w_s(\predP(x_1, ..., x_n))|\btheta)=\prl(\predP(x_1, ..., x_n)|s, \btheta).
\end{align}
Marginalize out all variables but $w_s(\predP(x_1, ..., x_n))$. $v(\predP(x_1, ..., x_n), \bw_s)$ is 1 if $w_s(\predP(x_1, ..., x_n))$ is, and 0 otherwise.

Next, assume $\phi = \neg \psi$. Then
\begin{align}
    &\sum_{\bw_s} p(\bw_s|\btheta) \cdot  v(\neg \psi, \bw_s) \\ 
    =& \sum_{\bw_s} p(\bw_s|\btheta) \cdot  (1 - v(\psi, \bw_s)) \\
    =& \sum_{\bw_s} p(\bw_s|\btheta) - \sum_{\bw_s} p(\bw_s|\btheta) \cdot v(\psi, \bw_s) \\
    =& 1 - \sum_{\bw_s} p(\bw_s|\btheta) \cdot v(\psi, \bw_s) = \prl(\neg \psi|s, \btheta)
\end{align}

Finally, assume $\varphi=\phi\wedge\psi$. Then
\begin{align}
    &\sum_{\bw_s} p(\bw_s|\btheta) \cdot  v(\phi \wedge\psi, \bw_s) \\ 
    =& \sum_{\bw_s} p(\bw_s|\btheta) \cdot v(\phi, \bw_s) \cdot v(\psi, \bw_s) \\ 
    =& \sum_{\bw_{\phi_s}} \sum_{\bw_{\psi_s}} p(\bw_{\phi_s}|\btheta) \cdot p(\bw_{\psi_s}|\btheta) \cdot v(\phi, \bw_{\phi_s}) \cdot v(\psi, \bw_{\psi_s})\cdot\\
    &\sum_{\bw_s\backslash \left(\bw_{\phi_s}\cup \bw_{\psi_s}\right)} p(\bw_s\backslash \left( \bw_{\phi_s}\cup \bw_{\psi_s}\right)|\btheta) \\ 
    =& \sum_{\bw_{\phi_s}} p(\bw_{\phi_s}|\btheta) \cdot v(\phi, \bw_{\phi_s}) \cdot\sum_{\bw_{\psi_s}} p(\bw_{\psi_s}|\btheta) \cdot v(\psi, \bw_{\psi_s}) \\ 
    =& \sum_{\bw_s} p(\bw_s|\btheta) \cdot v(\phi, \bw_s) \cdot\sum_{\bw_s} p(\bw_s|\btheta) \cdot v(\psi, \bw_s) \\ 
    =& \prl(\phi|s, \btheta)\cdot \prl(\psi|s, \btheta) = \prl(\phi\wedge\psi|s, \btheta)
\end{align}

Using this result and equations \ref{eq:deriv2} and \ref{eq:deriv3}, we find that
\begin{align}
    p(\corpus|D, \btheta) &=\prod_{\varphi\in\corpus} \sum_{\bw_\varphi} p(\bw_\varphi|\btheta) \cdot  v(\phi, \bw_\varphi) \\
    &= \prod_{\varphi\in\corpus} \prod_{s\in S_\varphi} \sum_{\bw_s} p(\bw_s|\btheta) \cdot v(\phi, \bw_s) \\
    &= \prod_{\varphi\in\corpus} \prod_{s\in S_\varphi} \prl(\phi|s, \btheta)
\end{align}

\end{appendices}
\end{document}